\documentclass[11pt,a4paper]{article}
\usepackage[hyperref]{emnlp-ijcnlp-2019}
\usepackage{times}
\usepackage{latexsym}
\usepackage{amssymb}
\usepackage{graphicx}
\usepackage{array, booktabs, bigdelim, makecell}

\usepackage{url}

\aclfinalcopy % Uncomment this line for the final submission

\title{Dual Attention Networks for Visual Reference Resolution in \\Visual Dialog}

\author{Gi-Cheon Kang\\
  Seoul National University \\
  {\tt chonkang@snu.ac.kr} \\\And
  Jaeseo Lim \\
  Seoul National University \\
  {\tt jaeseolim@snu.ac.kr} \\\And
  Byoung-Tak Zhang \\
  Seoul National University \\
  Surromind Robotics \\
  {\tt btzhang@snu.ac.kr} \\}
\date{}

\begin{document}
\maketitle
\begin{abstract}
  Visual dialog (VisDial) is a task which requires a dialog agent to answer a series of questions grounded in an image. Unlike in visual question answering (VQA), the series of questions should be able to capture a temporal context from a dialog history and utilizes visually-grounded information. Visual reference resolution is a problem that addresses these challenges, requiring the agent to resolve ambiguous references in a given question and to find the references in a given image. In this paper, we propose Dual Attention Networks (DAN) for visual reference resolution in VisDial. DAN consists of two kinds of attention modules, {\bf REFER} and {\bf FIND}. Specifically, REFER module learns latent relationships between a given question and a dialog history by employing a multi-head attention mechanism. FIND module takes image features and reference-aware representations (i.e., the output of REFER module) as input, and performs visual grounding via bottom-up attention mechanism. We qualitatively and quantitatively evaluate our model on VisDial v1.0 and v0.9 datasets, showing that DAN outperforms the previous state-of-the-art model by a significant margin.
\end{abstract}

\section{Introduction}

Thanks to the recent progresses in natural language processing and computer vision, there has been an extensive amount of effort towards developing a cognitive agent that jointly understand natural language and vision information. Over the last few years, vision-language tasks such as image captioning \cite{xu2015show} and visual question answering (VQA) \cite{antol2015vqa,Anderson2017up-down} have provided a testbed for developing a cognitive agent. However, the agent performing these tasks still has a long way to go to be used in real-world applications ({\it e.g}., aiding visually impaired users, interacting with humanoid robots) in that it does not consider the continuous interaction over time. Specifically, the interaction in image captioning is that the agent simply talks to human about visual content, without any input from human. While the VQA agent takes a question as input, it is required to answer a {\it single} question about a given image. 

Visual dialog (VisDial) \cite{das2017visual} task has been introduced as a generalized version of VQA. A dialog agent needs to answer a series of questions such as ``How many people are in the image?'', ``Are they indoors or outside?'', utilizing not only visually-grounded information but also contextual information from a dialog history. To address these two challenges, researchers have recently tackled a problem called visual reference resolution in VisDial. The problem of visual reference resolution is to resolve ambiguous expressions on their own ({\it e.g}., it, they, any other) and ground them to a given image. 

\begin{figure*}[ht!]
\label{figure:architecture1}
\centering
\includegraphics[height=7.8cm]{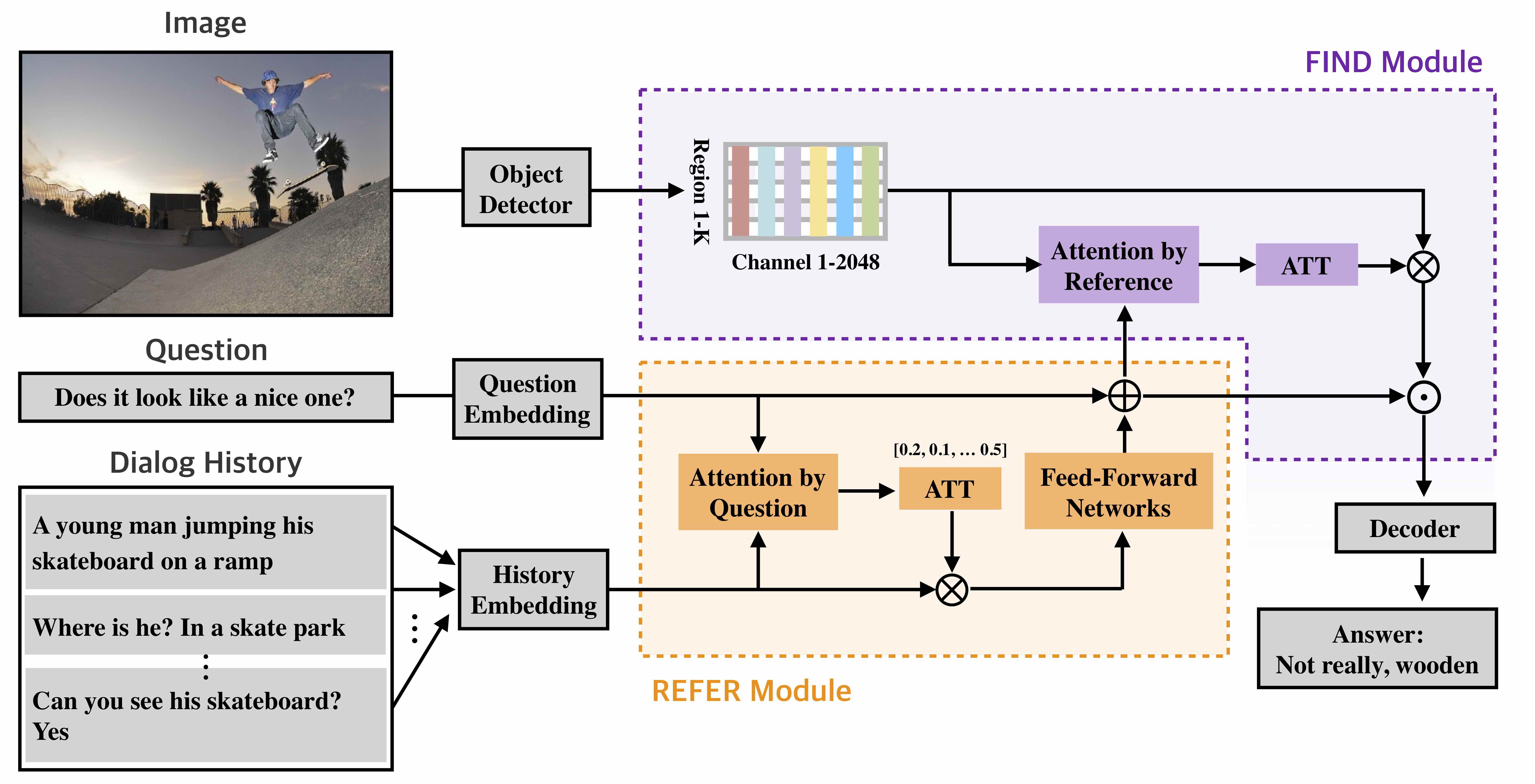}
\caption{An overview of Dual Attention Networks (DAN). We propose two kinds of attention modules, REFER and FIND. REFER learns latent relationships between a given question and a dialog history to retrieve the relevant previous dialogs. FIND performs visual grounding, taking image features and reference-aware representations ({\it i.e.,} the output of REFER). $\otimes$, $\oplus$, and $\odot$ denote matrix multiplication, concatenation and element-wise multiplication, respectively. The multi-layer perceptron is omitted in this figure for simplicity.}
\end{figure*}

In this paper, we address the visual reference resolution in a visual dialog task. We first hypothesize that humans address the visual reference resolution through a two-step process: (1) linguistically resolve the ambiguous questions by recalling the dialog history from one's memory and (2) find a local region of a given image for the resolved questions.  For example, as shown in Figure 1, the question ``Does it look like a nice one?'' is ambiguous on its own because we do not know what ``it'' refers to. So we believe that humans try to recall the dialog history and implicitly find out ``it'' refers to the ``skateboard''. After the resolution step, we believe that they will finally try to find the skateboard in the image and answer the question. For these processes, we propose Dual Attention Networks (DAN) which consists of two kinds of attention modules, REFER and FIND. REFER module learns to retrieve the relevant previous dialogs for clarifying ambiguous questions. Inspired by the self-attention mechanism \cite{vaswani2017attention}, REFER module computes multi-head attention over all previous dialogs in a sentence-level fashion, followed by feed-forward networks to get the {\it reference-aware} representations. FIND module takes image features and the {\it reference-aware} representations, and performs visual grounding via bottom-up attention mechanism. From this pipeline, we expect our proposed model to be capable of question disambiguation by using REFER module and ground the resolved reference properly to the given image. 

The main contributions of this paper are as follows. First, we propose Dual Attention Networks (DAN) for visual reference resolution in visual dialog based on REFER and FIND modules. Second, we validate our proposed model on the large-scale datasets: VisDial v1.0 and v0.9. Our model achieves a new state-of-the-art results compared to other methods. We also conduct ablation studies by four criteria to demonstrate the effectiveness of our proposed components. Third, we make a comparison between DAN and our baseline model to demonstrate the performance improvements on {\it semantically incomplete} questions needed to be clarified. Finally, we perform qualitative analysis of our model, showing that DAN reasonably attends to the dialog history and salient image regions. Our code is available at \url{https://github.com/gicheonkang/DAN-VisDial}. 

\section{Related Work}

\paragraph{Visual Dialog.} Visual dialog (VisDial) task was recently proposed by \cite{das2017visual}, providing a testbed for research on the interplay between computer vision and dialog systems. Accordingly, a dialog agent performing this task is not only required to find visual groundings of linguistic expressions but also capture semantic nuances from human conversation. Attention-based approaches were primarily proposed to address these challenges, including memory networks \cite{das2017visual}, history-conditioned image attentive encoder \cite{lu2017best}, sequential co-attention \cite{wu2018you},  and synergistic co-attention networks \cite{guo2019image}.

\paragraph{Visual Reference Resolution.} Recently, researchers have tackled a problem called visual reference resolution   \cite{seo2017visual,kottur2018visual,niu2018recursive} in VisDial. To resolve visual references, \cite{seo2017visual} proposed an attention memory which stores a sequence of previous visual attention maps in memory slots. They retrieved the previous visual attention maps by applying a soft attention over all the memory slots and combined it with a current visual attention. Furthermore, \cite{kottur2018visual} attempted to resolve visual references at a word-level, relying on an off-the-shelf parser. Similar to the attention memory \cite{seo2017visual}, they proposed a reference pool which stores visual attention maps of recognized entities and retrieved the weighted sum of the visual attention maps by applying a soft attention. \cite{niu2018recursive} proposed a recursive visual attention model that recursively reviews the previous dialogs and refines the current visual attention. The recursion is continued until the question itself is determined to be unambiguous. A binary decision whether the questions is ambiguous or not is made by Gumbel-Softmax approximation  \cite{jang2016categorical,maddison2016concrete}. To resolve the visual references, above approaches attempted to retrieve the visual attention of the previous dialogs, and applied it on the current visual attention. 
These approaches have limitations in that they store all previous visual attentions, while researches in human memory system show that the visual sensory-memory, due to its rapid decay property, hardly stores all previous visual attentions \cite{sperling1960information,sergent2011top}. Based on this biologically inspired motivation, our proposed model calculates the current visual attention by using linguistic cues ({\it i.e.,} dialog history). 

\begin{figure*}[ht!]
\label{figure:architecture2}
\centering
\includegraphics[height=4.4cm]{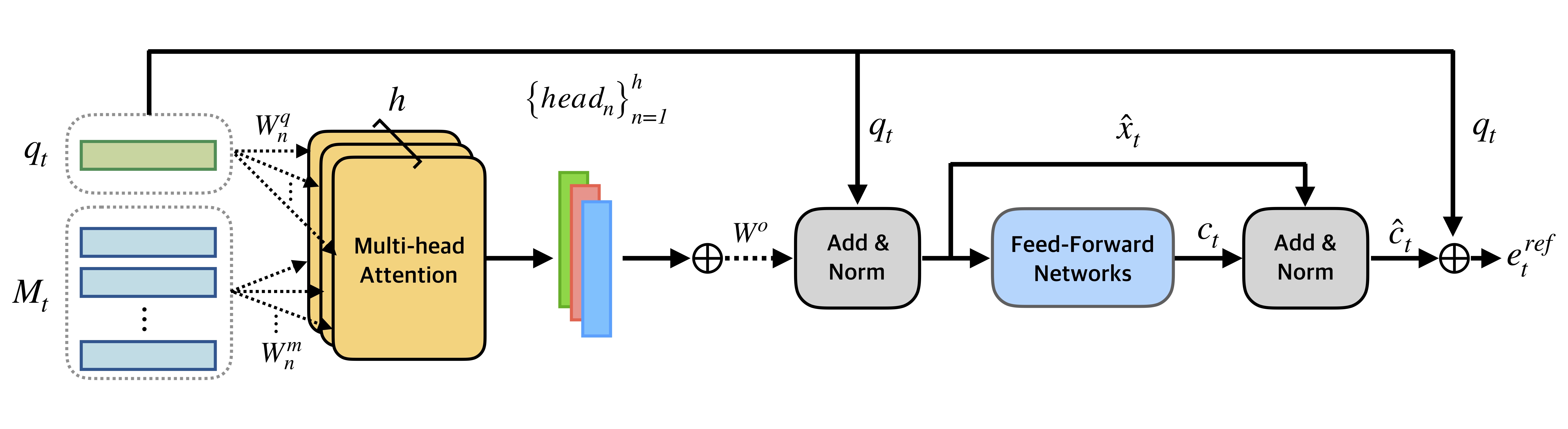}
\caption{Illustration of the single-layer REFER module. REFER module focuses on the latent relationship between the follow-up question and a dialog history to resolve ambiguous references in the question. We employ two submodule: multi-head attention and feed-forward networks. Multi-head attention computes the $h$ number of soft attentions over all elements of dialog history by using scaled dot product attention. Then, it returns the $h$ number of heads which are weighted by the attentions. Followed by the two-layer feed-forward networks, REFER module finally returns the reference-aware representations $e^{ref}_t$. $\oplus$ and Dotted line denote the concatenation operation and linear projection operation by the learnable matrices, respectively.}
\end{figure*}

\section{Proposed Algorithm}

In this section, we formally describe the visual dialog task and our proposed algorithm, Dual Attention Networks (DAN). The visual dialog task \cite{das2017visual} is defined as follows. A dialog agent is given an input such as an image {\bf{\it I}}, a follow-up question at round $\mathrm{t}$ as {\bf{\it $Q_t$}}, and a dialog history (including the image caption) until round $\mathrm{t-1}$,

{\bf{\it H}} = $($ $\underbrace{C}_{H_0}$, $\underbrace{(Q_1, A_1^{gt})}_{H_1}$, $\cdots$ , $\underbrace{(Q_{t-1}, A_{t-1}^{gt})}_{H_{t-1}}$$)$. \\ $A_t^{gt}$ denotes the ground truth answer ({\it i.e.,} human response) at round $\mathrm{t}$. By using these inputs, the agent is asked to rank a list of 100 candidate answers, {\it $A_t$ = $\left\{A_{t}^1, \cdots, A_{t}^{100} \right\}$}.

Given the problem setup, DAN for visual dialog task can be framed as an encoder-decoder architecture: (1) an encoder that jointly embeds the input ({\bf{\it I}}, {\bf{\it $Q_t$}}, {\bf{\it H}}) and (2) a decoder that converts the embedded representation into the ranked list $\hat{{\it A_t}}$. From this point of view, DAN consists of three components which are REFER, FIND, and the answer decoder. As shown in Figure 1, REFER module learns to attend relevant previous dialogs to resolve the ambiguous references in a given question {\bf{\it $Q_t$}}. FIND module learns to attend to the spatial image features that the output of REFER module describes. Answer decoder ranks the list of candidate answers ${\it A_t}$ given the output of FIND module.   

We first introduce the language features, as well as the image features in Sec. \hyperref[sec:inputrep]{3.1}. Then we describe the detailed architectures of the REFER and FIND modules in Sec. \hyperref[sec:refer]{3.2} and \hyperref[sec:find]{3.3}, respectively. Finally, we present the answer decoder in Sec. \hyperref[sec:ansdec]{3.4}.

\subsection{Input Representation}
\label{sec:inputrep}
\paragraph{Language Features.} We first embed each word in the follow-up question {\bf{\it $Q_t$}} to {\it $\left\{w_{t,1}, \cdots, w_{t,T} \right\}$} by using pre-trained GloVe \cite{pennington2014glove} embeddings, where {\it T} denotes the number of tokens in {\bf{\it $Q_t$}}. We then use a two-layer LSTM, generating a sequence of hidden states {\it $\left\{u_{t,1}, \cdots, u_{t,T} \right\}$}. Note that we use the last hidden state of the LSTM $u_{t,T}$ as a question feature, denoted as $q_{t} \in$ $\mathbb{R}^{L}$.  
\begin{equation}
    {u_{t,i}} = \mathrm{LSTM(} {w_{t,i}}, {u_{t, i-1}} \mathrm{)}
\end{equation} 
\begin{equation}
    {q_{t}} = {u_{t,T}}
\end{equation} 
Also, each element in the dialog history {\it $\left\{H_{i} \right\}$$_{i=0}^{t-1}$} and the candidate answers {\it $\left\{A_{t}^i \right\}$$_{i=1}^{100}$} are embedded as the follow-up question, yielding {\it $\left\{h_{i} \right\}$$_{i=0}^{t-1}$} $\in$ $\mathbb{R}^{t \times L}$  and {\it $\left\{o_{t}^i \right\}$$_{i=1}^{100}$} $\in$ $\mathbb{R}^{100 \times L}$. {\bf{\it $Q_t$}}, {\bf{\it H}}, and {\bf{\it $A_t$}} are embedded with same word embedding vector and three different LSTMs.

\paragraph{Image Features.} Inspired by bottom-up attention \cite{Anderson2017up-down}, we use the Faster R-CNN \cite{ren2015faster} pre-trained with Visual Genome \cite{krishna2017visual} to extract the object-level image features. We denote the output features as $ v \in$ $\mathbb{R}^{K \times V}$, where {\it K} and {\it V} are the total number of object detection features per image and dimension of the each feature, respectively. We adaptively extract the number of object features {\it K} ranging from 10 to 100 for reflecting the complexity of each image. {\it K} is fixed during training.

\subsection{REFER Module}
\label{sec:refer}
In this section, we formally describe the single-layer REFER module. Given the question and dialog history features, REFER module aims to attend to the 
most relevant elements of dialog history with respect to the given question. Specifically, we first compute scaled dot product attention \cite{vaswani2017attention} in multi-head settings which are called multi-head attention. Let $q_{t}$ and $M_{t}$ = {\it $\left\{h_{i} \right\}_{i=0}^{t-1}$} be the question and dialog history feature vectors, respectively.  $q_{t}$ and $M_{t}$ are projected to $d_{ref}$ dimensions by different and learnable projection matrices. We then conduct dot product of these two projected matrices, divide by $\sqrt{d_{ref}}$, and apply the softmax function to obtain the attention weights on the all elements in the dialog history. It is formulated as below,
\begin{equation}
    head_n \mathrm{\, = Attention (} {q_{t}}W_n^q, {M_{t}}W_n^m \mathrm{)} 
\end{equation}
\begin{equation}
    \mathrm{\; Attention(} {a, b}  \mathrm{) = softmax(} { \frac{{ab}^\top}{\sqrt{{d_{ref}}}} \mathrm{)}}{b}
\end{equation}
where $W_n^q \in$ $\mathbb{R}^{L \times d_{ref}}$ and $W_n^m \in$ $\mathbb{R}^{L \times d_{ref}}$.  Note that dot product attention is computed $h$ times with different projection matrices, yielding {\it $\left\{head_n \right\}_{n=1}^{h}$}. Accordingly, we can get the multi-head representations $x_t$, concatenating all {\it $\left\{head_n \right\}_{n=1}^{h}$}, followed by linear projection. Also, we can compute $\hat{x}_t$ by applying a residual connection \cite{he2016deep}, followed by layer normalization \cite{ba2016layer}.
\begin{equation}
    {x_t} = \left( head_1 \oplus \cdots \oplus head_h \right)W^o  
\end{equation}
\begin{equation}
    {\hat{x}_t} = \mathrm{LayerNorm(} {x_t} + q_t \mathrm{)} 
\end{equation}
where $\oplus$ denotes the concatenation operation, and $W^o \in$ $\mathbb{R}^{hd_{ref} \times L}$ is the projection matrix. Next, we apply $\hat{x}_t$ to two-layer feed-forward networks with a ReLU in between, where $W_1^f \in$ $\mathbb{R}^{L \times 2L}$ and $W_2^f \in$ $\mathbb{R}^{2L \times L}$. The residual connection and layer normalization is also applied in this step.

\begin{equation}
    {c_t} = \mathrm{ReLU(} {\hat{x}_t}W_1^f + b_1^f \mathrm{)}W_2^f + b_2^f
\end{equation}
\begin{equation}
    {\hat{c}_t} = \mathrm{LayerNorm(} {c_t} + {\hat{x}_t} \mathrm{)} 
\end{equation}
\begin{equation}
    {e_t^{ref}} = {\hat{c}_t} \oplus {q_t} 
\end{equation}
Finally, REFER module returns the {\it reference-aware} representations by concatenating the contextual representation $\hat{c}_t$ and the original question representation $q_t$, denoted as $e^{ref}_t \in$ $\mathbb{R}^{2L}$. In this work, we use $d_{ref}=256$. Figure 2 illustrates the pipeline of the REFER module.

Furthermore, we stack the REFER modules in multiple layers to get a high-level abstraction of the reference-aware representations. Details are to be discussed in Sec. \hyperref[sec:abl]{4.5}. 

\subsection{FIND Module}
\label{sec:find}
Instead of relying on the visual attention maps of the previous dialogs as in \cite{seo2017visual,kottur2018visual,niu2018recursive}, we expect the FIND module to attend to the most relevant regions of the image with respect to the reference-aware representations ({\it i.e.,} the output of REFER module). In order to implement the visual grounding for the reference-aware representations, we take inspiration from bottom-up attention mechanism \cite{Anderson2017up-down}. Let $v$ $\in$ $\mathbb{R}^{K \times V}$ and $e^{ref}_t$ $\in$ $\mathbb{R}^{2L}$ be the image feature vectors and reference-aware representations, respectively. We first project these two vectors to $d_{find}$ dimensions and compute soft attention over all the object detection features as follows: \\

\begin{equation}
    {r_t} = f_{v}({v}) \odot f_{ref}({e^{ref}_t})  
\end{equation}
\begin{equation}
    \alpha_t = \mathrm{softmax(} {r_t}W^{r} + b^{r} \mathrm{)}  
\end{equation}
where $f_{v}(\cdot)$ and $f_{ref}(\cdot)$ denote the two-layer multi-layer perceptrons which convert to $d_{find}$ dimensions, and $W^r \in$ $\mathbb{R}^{d_{find} \times 1}$ is the projection matrix for the softmax activation. $\odot$ denotes hadamard product ({\it i.e.,} element-wise multiplication). From these equations, we can get the visual attention weights $\alpha_t \in$ $\mathbb{R}^{K \times 1}$. Next, we apply the visual attention weights to $v$ and compute the vision-language joint representations as follows:
\begin{equation}
    {\hat{v}_t} = \sum_{j=1}^K \alpha_{t, j} {v_j}  
\end{equation}
\begin{equation}
    {z_t} = f_{v}^{\prime}({\hat{v}_t}) \odot f_{ref}^{\prime}({e^{ref}_t})  
\end{equation}
\begin{equation}
    {e^{find}_t} = {z_t}W^{z} + b^{z}  
\end{equation}
where $f_{v}^{\prime}(\cdot)$ and $f_{ref}^{\prime}(\cdot)$ also denote the two-layer multi-layer perceptrons which convert to $d_{find}$ dimensions, and $W^z \in$ $\mathbb{R}^{d_{find} \times L}$ is the projection matrix. Note that $e^{find}_t \in$ $\mathbb{R}^{L}$ is the output representations of the encoder as well as FIND module which is decoded to score the list of candidate answers. In this work, we use $d_{find}=1024$. 

\subsection{Answer Decoder}
\label{sec:ansdec}
Answer decoder computes each score of candidate answers via a dot product with the embedded representation $e^{find}_t$, followed by a softmax activation to get a categorical distribution over the candidates. Let $O_t$ = {\it $\left\{o_{t}^i \right\}$$_{i=1}^{100}$} $\in$ $\mathbb{R}^{100 \times L}$ be the feature vectors of 100 candidate answers. The distribution $p_t$ is formulated as follows:
\begin{equation}
    {p_t} = \mathrm{softmax(} {e^{find}_t} {O_t}^\top \mathrm{)}  
\end{equation}
In training phase, DAN is optimized by minimizing the cross-entropy loss between the one-hot encoded label vector ({\it{i.e}}., $y_t$) and probability distribution ({\it{i.e}}., $p_t$).
\begin{equation}
    \mathcal{L}(\theta) = -\sum_{k} {y_{t, k}} \, \mathrm{log \,} {p_{t, k}}  
\end{equation}
Where $p_{t,k}$ denotes the probability of the $k$-$th$ candidate answer at round $\mathrm{t}$. In test phase, the list of candidate answers is ranked by the distribution $p_t$, and evaluated by the given metrics.

\begin{table*}[ht!]
\centering
\resizebox{\textwidth}{!}{
\begin{tabular}{lcccccccccccc}
\hline
\toprule
\multirow{2}{*}{} & \multicolumn{6}{c}{VisDial v1.0 (test-std)}  & \multicolumn{5}{c}{VisDial v0.9 (val)} \\ 
\cmidrule(lr){2-7}\cmidrule(lr){8-12}
 & NDCG & MRR & R@1 & R@5 & R@10 & Mean & MRR & R@1 & R@5 & R@10 & Mean\\
\midrule
LF \cite{das2017visual} & 45.31 & 55.42 & 40.95 & 72.45 & 82.83 & 5.95 & 58.07 & 43.82 & 74.68 & 84.07 & 5.78\\
HRE \cite{das2017visual} & 45.46 & 54.16 & 39.93 & 70.45 & 81.50 & 6.41 & 58.46 & 44.67 & 74.50 & 4.22 & 5.72\\
MN \cite{das2017visual} & 47.50 & 55.49  & 40.98 & 72.30 & 83.30 & 5.92 & 59.65 & 45.55 & 76.22 & 85.37 & 5.46\\
HCIAE \cite{lu2017best} & - & - & - & - & - & - & 62.22 & 48.48 & 78.75 & 87.59 & 4.81 \\
AMEM \cite{seo2017visual}  & - & - & - & - & - & - & 62.27 & 48.53 & 78.66 & 87.43 & 4.86 \\
CoAtt \cite{wu2018you} & - & - & - & - & - & - & 63.98 & 50.29 & 80.71 & 88.81 & 4.47 \\
CorefNMN \cite{kottur2018visual} & 54.70 & 61.50  & 47.55 & 78.10 & 88.80 & 4.40 & 64.10 & 50.92 & 80.18 & 88.81 & 4.45\\
RvA \cite{niu2018recursive} & 55.59 & 63.03  & 49.03 & 80.40 & 89.83 & 4.18 & 66.34 & 52.71 & {\bf 82.97} & {\bf 90.73} & {\bf 3.93} \\
Synergistic \cite{guo2019image} & 57.32 & 62.20 & 47.90 & {\bf 80.43} & {\bf 89.95} & {\bf 4.17} & - & - & - & - & - \\ 
\midrule
DAN (ours) & {\bf 57.59} & {\bf 63.20}  & {\bf 49.63} & 79.75 & 89.35 & 4.30 & {\bf 66.38} & {\bf 53.33} & 82.42 & 90.38 & 4.04 \\
\bottomrule
\hline
\end{tabular}
}
\caption{Retrieval performance on VisDial v1.0 and v0.9 datasets, measured by normalized discounted cumulative gain (NDCG), mean reciprocal rank (MRR), recall @k (R@k), and mean rank. The higher the better for NDCG, MRR, and R@k, while the lower the better for mean rank. DAN outperforms all other models across NDCG, MRR, and R@1 on both datasets. NDCG is not supported in v0.9 dataset.}
\label{tab:plain}
\end{table*}

\section{Experiments}
In this section, we describe the details of our experiments on the VisDial v1.0 and v0.9 datasets. We first introduce the VisDial datasets, evaluation metrics, and implementation details in Sec. \hyperref[sec:dset]{4.1}, Sec. \hyperref[sec:eval]{4.2}, and Sec. \hyperref[sec:impdet]{4.3}, respectively. Then we report the quantitative results by comparing our proposed model with the state-of-the-art approaches and baseline model in Sec. \hyperref[sec:quan]{4.4}. Then, we conduct the ablation studies by four criteria to report the relative contributions of each components in Sec. \hyperref[sec:abl]{4.5}. Finally, we provide the qualitative results in Sec. \hyperref[sec:qual]{4.6}.

\subsection{Datasets}
\label{sec:dset}
We evaluate our proposed model on the VisDial v0.9 and v1.0 dataset. VisDial v0.9 dataset \cite{das2017visual} has been collected from two annotators chatting log about MS-COCO \cite{lin2014microsoft} images. Each dialog is made up of an image, a caption from MS-COCO dataset and 10 QA pairs. As a result, VisDial v0.9 dataset contains 83k dialogs and 40k dialogs as train and validation splits, respectively. Recently, VisDial v1.0 dataset \cite{das2017visual} has been released with an additional 10k COCO-like images from Flickr. Dialogs for the additional images have been collected similar to v0.9. Overall, VisDial v1.0 dataset contains 123k (all dialogs from v0.9), 2k, and 8k dialogs as train, validation, and test splits, respectively.

\subsection{Evaluation Metrics}
\label{sec:eval}
We evaluate individual responses at each question in a retrieval setting according to \cite{das2017visual}. Specifically, the dialog agent is given a list of 100 candidate answers of each question and asked to rank the list. There are three kinds of evaluation metrics for retrieval performance: (1) mean rank of human response, (2) recall@k ({\it i.e.,} existence of the human response in top-k ranked response), and (3) mean reciprocal rank (MRR). Mean rank, recall@k, and MRR are highly correlated with the rank of human response. In addition, \cite{das2017visual} proposed to use the robust evaluation metric, normalized discounted cumulative gain (NDCG). NDCG takes into account all relevant answers from the ranked list, where the relevance scores are densely annotated for VisDial v1.0 test split. NDCG penalizes the lower rank of the candidate answers with high relevance scores.

\subsection{Implementation Details}
\label{sec:impdet}
The dimension of image features $V$ and hidden states in all LSTM $L$ is 2048 and 512, respectively. All the language intputs are embedded into a 300-dimensional vector initialized by GloVe \cite{pennington2014glove}. The number of attention heads $h$ is fixed to 4 except for the ablation study that changes it. We apply Adam optimizer \cite{kingma2014adam} with learning rate 1 $\times 10^{-3}$, decreased by 1 $\times 10^{-4}$ per epoch until epoch 7, decayed by 0.5 per epoch from 8 to 12 epochs.

\subsection{Quantitative Results}
\label{sec:quan}
\paragraph{Compared Methods.} We compare our proposed model with the state-of-the-art approaches on VisDial v1.0 and v0.9 datasets, which can be categorized into three groups: (1) Fusion-based approaches (LF and HRE \cite{das2017visual}), (2) Attention-based approaches (MN \cite{das2017visual}, HCIAE \cite{lu2017best}, CoAtt \cite{wu2018you} and Synergistic \cite{guo2019image}), and (3) Approaches that deal with visual reference resolution in VisDial (AMEM \cite{seo2017visual}, CorefNMN \cite{kottur2018visual} and RvA \cite{niu2018recursive}). Our proposed model belongs to the third category.

\paragraph{Results on VisDial v1.0 and v0.9 datasets.} As shown in Table 1, DAN significantly outperforms all other approaches on NDCG, MRR, and R@1, including the previous state-of-the-art method, Synergistic \cite{guo2019image}. Specifically, DAN improves approximately 1.0\% on MRR, 1.7\% on R@1 and 0.3\% on NDCG in VisDial v1.0 dataset. The results indicate that our proposed model ranks higher than all other methods on both single ground-truth answer (R@1) and all relevant answers on average (NDCG).

\begin{table}
\centering
\resizebox{\columnwidth}{!}{
\begin{tabular}{lcccccc}
\hline
\toprule
Model  & NDCG & MRR & R@1 & R@5 & R@10 & Mean \\
\midrule
MS ConvAI & 55.35 & 63.27 & 49.53 & 80.40 & 89.60 & 4.15 \\
USTC-YTH & 56.47 & 61.44 & 47.65 & 78.13 & 87.88 & 4.65 \\
Synergistic & 57.88 & 63.42 & 49.30 & 80.77 & 90.68 & 3.97 \\
\midrule
DAN (ours) & {\bf 59.36} & {\bf 64.92} & {\bf 51.28} & {\bf 81.60} & {\bf 90.88} & {\bf 3.92}\\
\bottomrule
\hline
\end{tabular}
}
\caption{Test-std performance of ensemble model on VisDial v1.0 dataset. We cite top-three entries from VisDial Challenge 2018 Leaderboard.}
\label{tab:t3}
\end{table}

\paragraph{Results on ensemble model.} We report the performance of ensemble model in comparison with the top-three entries in the leaderboard\footnote{\url{https://evalai.cloudcv.org/web/challenges/challenge-page/103/leaderboard}} of VisDial Challenge 2018. We ensemble six DAN models, using the number of attention heads ({\it i.e.,} $h$) ranging from one to six. We average the probability distribution ({\it i.e.,} $p_t$) of the six models to rank the candidate answers. In Table 2, our model significantly outperforms all three challenge entries, including the challenge winner model, Synergistic \cite{guo2019image}. They ensembled ten models with different weight initialization and also used bottom-up attention features \cite{Anderson2017up-down} as image features. 

\paragraph{Results on semantically complete \& incomplete questions.} We first define the questions that contain one or more pronouns ({\it i.e.,} it, its, they, their, them, these, those, this, that, he, his, him, she, her) as the {\it semantically incomplete} (SI) questions. Also, we can declare the questions that do not have pronouns as {\it semantically complete} (SC) questions. Then, we have checked the contribution of the {\it{reference-aware}} representations for the SC and SI questions, respectively. Specifically, we make a comparison between DAN, which utilizes {\it reference-aware} representations ({\it i.e.,} $e^{ref}_t$), and No REFER, which exploits question representations ({\it i.e.,} $q_t$) only. From the Table 3, we draw three observations: (1) DAN shows significantly better results than the No REFER model for SC questions. It validates that the context from dialog history enriches the question information, even when the question is semantically complete. (2) SI questions obtain more benefits from the dialog history than SC questions. It indicates that DAN is more robust to the SI questions than SC questions. (3) A dialog agent faces greater difficulty in answering SI questions compared to SC questions. No REFER is equivalent to the FIND + RPN model in the ablation study section. 
\begin{table}
\centering
\resizebox{\columnwidth}{!}{
\begin{tabular}{lccccc}
\hline
\toprule
Model  & MRR & R@1 & R@5 & R@10 & Mean\\
\midrule
\ldelim\{{3}{*}[\rotatebox{90}{SC}] No REFER & 61.85 & 47.80 & 79.10 & 88.43 & 4.49 \\
\hspace{1.7em} DAN & 64.81 & 51.22 & 81.63 & 90.19 & 4.03\\
\hspace{1.7em} Improvements & 2.96 & 3.42 & 2.53 & 1.76 & 0.46\\
\midrule
\ldelim\{{3}{*}[\rotatebox{90}{SI}] No REFER & 58.44 & 44.38 & 75.36 & 85.48 & 5.36 \\
\hspace{1.7em} DAN & 61.77 & 48.13 & 78.43 & 87.81 & 4.70\\
\hspace{1.7em} Improvements & 3.33 & 3.75 & 3.07 & 2.33 & 0.66\\
\bottomrule
\hline
\end{tabular}
}
\caption{VisDial v1.0 validation performance on the semantically complete (SC) and incomplete (SI) questions. We observe that SI questions obtain more benefits from the dialog history than SC questions.}
\label{tab:t4}
\end{table}

\subsection{Ablation Study}
\label{sec:abl}
In this section, we perform ablation study on VisDial v1.0 validation split with the following four model variants: (1) Model only using the single attention module, (2) Model that uses different image features (pre-trained VGG-16 is used), (3) Model that does not use the residual connection in REFER module, and (4) Model that stacks the REFER modules up to four layers with each different number of attention heads.
\begin{table}
\centering
\resizebox{0.9\columnwidth}{!}{
\begin{tabular}{lccccc}
\hline
\toprule
Model  & MRR Score \\
\midrule
FIND   & 57.85 \\  
FIND + RPN & 60.80 \\ 
REFER    & 57.18  \\
REFER + Res & 58.69 \\
\midrule
REFER + FIND & 60.98 \\ 
REFER + Res + FIND & 61.86 \\
REFER + FIND + RPN & 63.47 \\
REFER + Res + FIND + RPN & 63.88\\
%REFER ($2$) + Res + FIND + RPN & 64.17 \\
\bottomrule
\hline
\end{tabular}
}
\caption{Ablation studies on VisDial v1.0 validation split. 
Res and RPN denote the residual connection and the region proposal networks, respectively.}
\label{tab:t5}
\end{table}

\begin{figure}[ht]
\label{figure:architecture4}
\centering
\includegraphics[height=5.6cm, width=7.9cm]{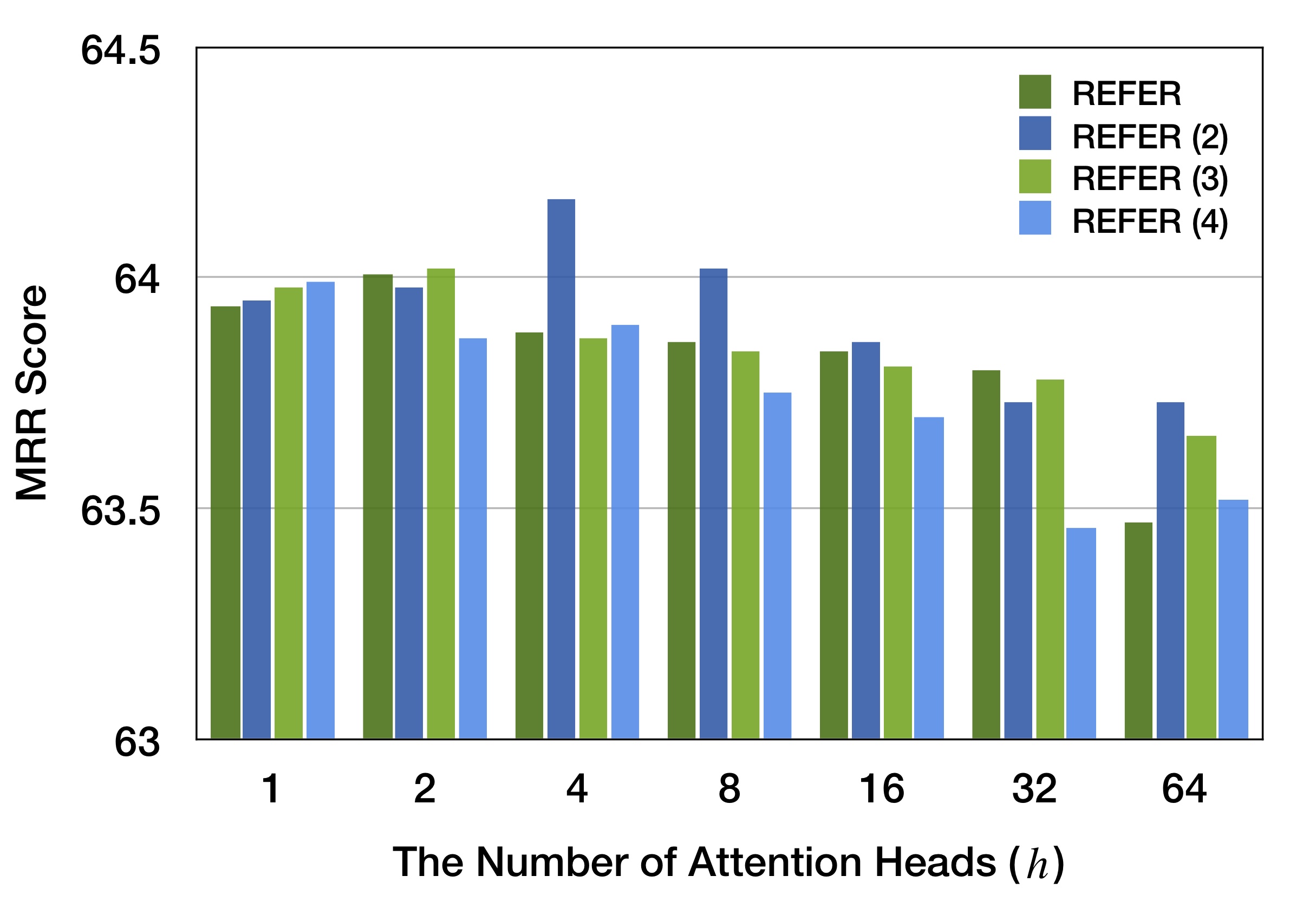}
\caption{Ablation study on a different number of attention heads and REFER stacks. REFER ($n$) indicates that DAN uses a stack of $n$ identical REFER modules.}
\end{figure}

\begin{figure*}[ht!]
\label{figure:architecture3}
\centering
\includegraphics[height=7.5cm, width=15cm]{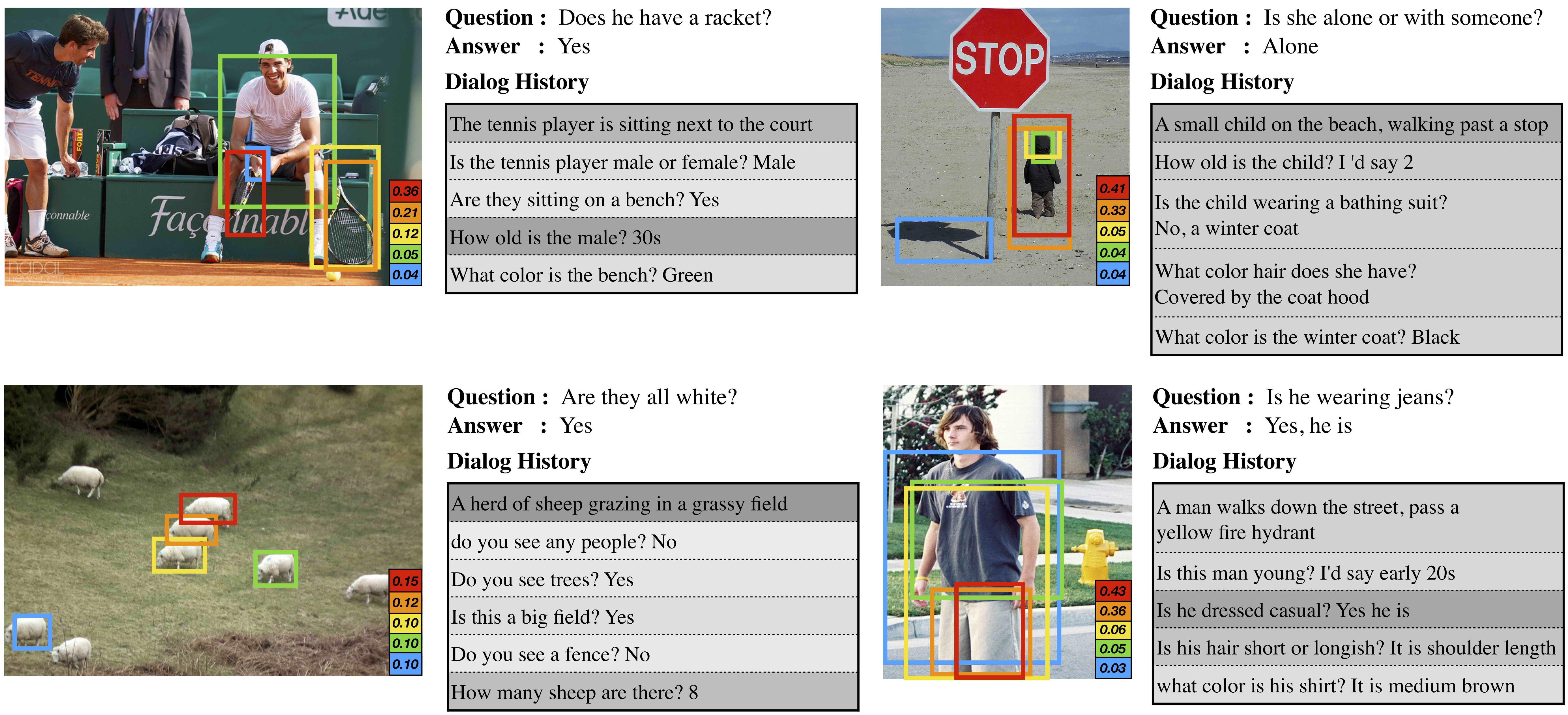}
\caption{Qualitative results on the VisDial v1.0 dataset. We visualize the attention over dialog history from REFER module and the visual attention from FIND module. The object detection features with top five attention weights are marked with colored box. A red colored box indicates the most salient visual feature.  Also, the attention from REFER module is represented as shading, darker shading indicates the larger attention weight for each element of the dialog history. Our proposed model not only responds to the correct answer, but also selectively pays attention to the previous dialogs and salient image regions.}
\end{figure*}

\paragraph{Single Module.}
The first four rows in Table 4 show the performance of a single module. FIND denotes the use of FIND module only, and REFER denotes the use of single-layer REFER module only. Specifically, REFER uses the output of REFER module as the encoder outputs. On the other hand, FIND does not take the reference-aware representations ({\it i.e.,} $e^{ref}_t$) but the question feature ({\it i.e.,} $q_t$). The single models show relatively poor performance compared with the dual module model. We believe that the results validate two hypotheses: (1) VisDial task requires contextual information from dialog history as well as the visually-grounded information. (2) REFER and FIND modules have complementary modeling abilities. 
\paragraph{Image Features in FIND Module.} To report the impact of image features, we replace the bottom-up attention features \cite{Anderson2017up-down} with ImageNet pre-trained VGG-16 \cite{simonyan2014very} features. In detail, we use the output of the VGG-16 {\it pool5} layer as image features. In Table 4, RPN denotes the use of the region proposal networks \cite{ren2015faster} which are equivalent to the use of bottom-up attention features. Similar to VQA task, we observe that DAN with bottom-up attention features achieves better performance than with VGG-16 features. In other words, the use of object-level features boosts the MRR performance of DAN.
\paragraph{Residual Connection in REFER Module.}
We also conduct an ablation study to investigate the effectiveness of the residual connection in REFER module. As shown in Table 4, the use of the residual connection ({\it i.e.,} Res) boosts the MRR score of DAN. In other words, DAN utilizes the excellence of deep residual learning as in \cite{he2016deep, rocktaschel2015reasoning, yang2016stacked, kim2016multimodal,  vaswani2017attention}.  
\paragraph{Stack of REFER Modules \& Attention Heads.} 

We stack the REFER modules up to four layers with each different number of attention heads, $h \in \left\{1, 2, 4, 8, 16, 32, 64 \right\}$. In other words, we conduct the ablation experiments with twenty-eight models to set the hyperparameters of our model. Figure 3 shows the results of the ablation experiments. For $n \ge 2$, REFER ($n$) indicates that DAN uses a stack of $n$ identical REFER modules. Specifically, for each pair of successive modules, the output of the previous REFER module is fed into the next REFER module as a query ({\it i.e.,} $q_t$). Due to the small number of elements in each dialog history, the overall performance pattern shows a tendency to decrease as the number of attention heads increases. It turns out that the two-layer REFER module with four attention heads ({\it i.e.,} REFER (2) and $h=4$) performs the best among all models in ablation study, recording 64.17\% on MRR. 

\subsection{Qualitative Results}
\label{sec:qual}
In this section, we visualize the inference mechanism of our proposed model. Figure 4 shows the qualitative results of DAN. 
Given a question that is needed to be clarified, DAN correctly answers the question by selectively attending to each element of the dialog history and salient image regions. In case of the visual attention, we mark the object detection features with top five attention weights of each image. On the other hand, the attention weights from REFER module are represented as shading; darker shading indicates the larger attention weight for each element of the dialog history. These attention weights are calculated by averaging over all the attention heads. 

\section{Conclusion}
We introduce Dual Attention Networks (DAN) for visual reference resolution in visual dialog task. DAN explicitly divides the visual reference resolution problem into a two-step process. Rather than relying on the previous visual attention maps as in prior works, DAN first linguistically resolves ambiguous references in a given question by using REFER module. Then, it grounds the resolved references in the image by using FIND module. We empirically validate our proposed model on VisDial v1.0 and v0.9 datasets. DAN achieves the new state-of-the-art performance, while being simpler and more grounded.

\paragraph{Acknowledgements}
The authors would like to thank Woosuk Choi, Seunghee Yang, Junseok Park, and Delilah Hollweg for helpful comments and editing. This work was partly supported by the Korea government (2015-0-00310-SW.StarLab, 2017-0-01772-VTT, 2018-0-00622-RMI, 2019-0-01367-BabyMind, 10060086-RISF, P0006720-GENKO), and the ICT at Seoul National University.

\bibliography{emnlp-ijcnlp-2019}
\bibliographystyle{acl_natbib}
\end{document}